%% file: main.tex
\input{_constants}
\arxiv 
\pdfoutput=1
\documentclass[10pt,twocolumn,letterpaper]{article}
\input{cvpr_header}
\begin{document}
\title{\paperTitle}
\author{\authorBlock}
\maketitle

\input{00_abstract}

\input{01_intro}

\input{02_related}

\input{03_method}

\input{04_experiments}

\input{10_conclusion}

\ifarxiv \input{12_appendix} \fi

{\small
\bibliographystyle{ieee_fullname}
\bibliography{11_references}
}

\end{document}

%% file: _constants.tex
\def\cvprPaperID{1147}
\def\confName{CVPR}
\def\confYear{2023}

\def\paperTitle{Aligning Bag of Regions for Open-Vocabulary Object Detection}

\def\authorBlock{
Size Wu$^{1}$ \quad Wenwei Zhang$^{1}$ \quad Sheng Jin$^{2,3}$ \quad Wentao Liu$^{3,4}$ \quad Chen Change Loy$^{1}$\thanks{Corresponding author.}  \\
$^{1}$Nanyang Technological University \quad $^{2}$ The University of Hong Kong \\ 
$^{3}$ SenseTime Research and Tetras.AI \quad $^{4}$ Shanghai AI Laboratory \\
{\tt\small\{size001, wenwei001,  ccloy\}@ntu.edu.sg} \quad \tt\small\{jinsheng, liuwentao\}@sensetime.com \\ 
}

\newif\ifreview 
\newif\ifarxiv \newcommand{\arxiv}{\arxivtrue}
\newif\ifcamera 
\newif\ifrebuttal 
\def\method{\text{BARON}}


%% file: cvpr_header.tex
\ifreview \usepackage[review]{cvpr} \fi
\ifarxiv \usepackage[pagenumbers]{cvpr} \fi
\ifrebuttal \usepackage[rebuttal]{cvpr} \fi
\ifcamera \usepackage{cvpr} \fi

\usepackage{graphicx}
\usepackage{amsmath}
\usepackage{amssymb}
\usepackage{booktabs}

\input{_macros}  

\usepackage{xr-hyper}

\makeatletter
\newcommand*{\addFileDependency}[1]{
  \typeout{(#1)}
  \@addtofilelist{#1}
  \IfFileExists{#1}{}{\typeout{No file #1.}}
}

\makeatother

\usepackage[pagebackref,breaklinks,colorlinks]{hyperref}
\usepackage[capitalize]{cleveref}
\crefname{section}{Sec.}{Secs.}
\crefname{table}{Table}{Tables}
\crefname{figure}{Fig.}{Figs.}

%% file: _macros.tex
\usepackage{times}
\usepackage{microtype}
\usepackage{epsfig}
\usepackage[table,xcdraw]{xcolor}
\usepackage{caption}
\usepackage{float}
\usepackage{placeins}
\usepackage{color, colortbl}
\usepackage{stfloats}
\usepackage{enumitem}
\usepackage{tabularx}
\usepackage{xstring}
\usepackage{multirow}
\usepackage{xspace}
\usepackage{url}
\usepackage{subcaption}
\usepackage{xcolor}
\usepackage[hang,flushmargin]{footmisc}

\definecolor{citecolor}{HTML}{0071bc}
\definecolor{ourscolor}{HTML}{c2d1e5}
\usepackage{colortbl}  
\usepackage[pagebackref,breaklinks,colorlinks,citecolor=citecolor]{hyperref}

\newcommand{\myparagraph}[1]{{\noindent\bf #1}}
\newlength\savewidth\newcommand\shline{\noalign{\global\savewidth\arrayrulewidth
  \global\arrayrulewidth 1pt}\hline\noalign{\global\arrayrulewidth\savewidth}}
\newcommand{\tablestyle}[2]{\setlength{\tabcolsep}{#1}\renewcommand{\arraystretch}{#2}\centering\small}

\ifcamera \usepackage[accsupp]{axessibility} \fi

% ...

\ifarxiv  \fi

\newcommand{\R}[1]{{%
    \textbf{%
        \ifstrequal{#1}{1}{\textcolor{red}{R#1}}{%
        \ifstrequal{#1}{2}{\textcolor{blue}{R#1}}{%
        \ifstrequal{#1}{3}{\textcolor{magenta}{R#1}}{%
        \ifstrequal{#1}{4}{\textcolor{teal}{R#1}}{%
                           \textcolor{cyan}{R#1}%
        }}}}%
    }%
}}

%% file: 00_abstract.tex
\begin{abstract}
Pre-trained vision-language models (VLMs) learn to align vision and language representations on large-scale datasets, where each image-text pair usually contains a bag of semantic concepts. 
However, existing open-vocabulary object detectors only align region embeddings \emph{individually} with the corresponding features extracted from the VLMs.
Such a design leaves the compositional structure of semantic concepts in a scene under-exploited, although the structure may be implicitly learned by the VLMs.
In this work, we propose to align the embedding of \emph{bag of regions} beyond individual regions.
The proposed method groups contextually interrelated regions as a bag.
The embeddings of regions in a bag are treated as embeddings of words in a sentence, and they are sent to the text encoder of a VLM to obtain the bag-of-regions embedding, which is learned to be aligned to the corresponding features extracted by a frozen VLM.
Applied to the commonly used Faster R-CNN, our approach surpasses the previous best results by 4.6 box AP$_\text{50}$ and 2.8 mask AP on novel categories of open-vocabulary COCO and LVIS benchmarks, respectively. 
Code and models are available at \url{https://github.com/wusize/ovdet}.

\vspace{-12pt}
\end{abstract}

%% file: 01_intro.tex
\section{Introduction}
\label{sec:intro}

\begin{figure}[t]
	\centering
	\includegraphics[width=0.5\textwidth]{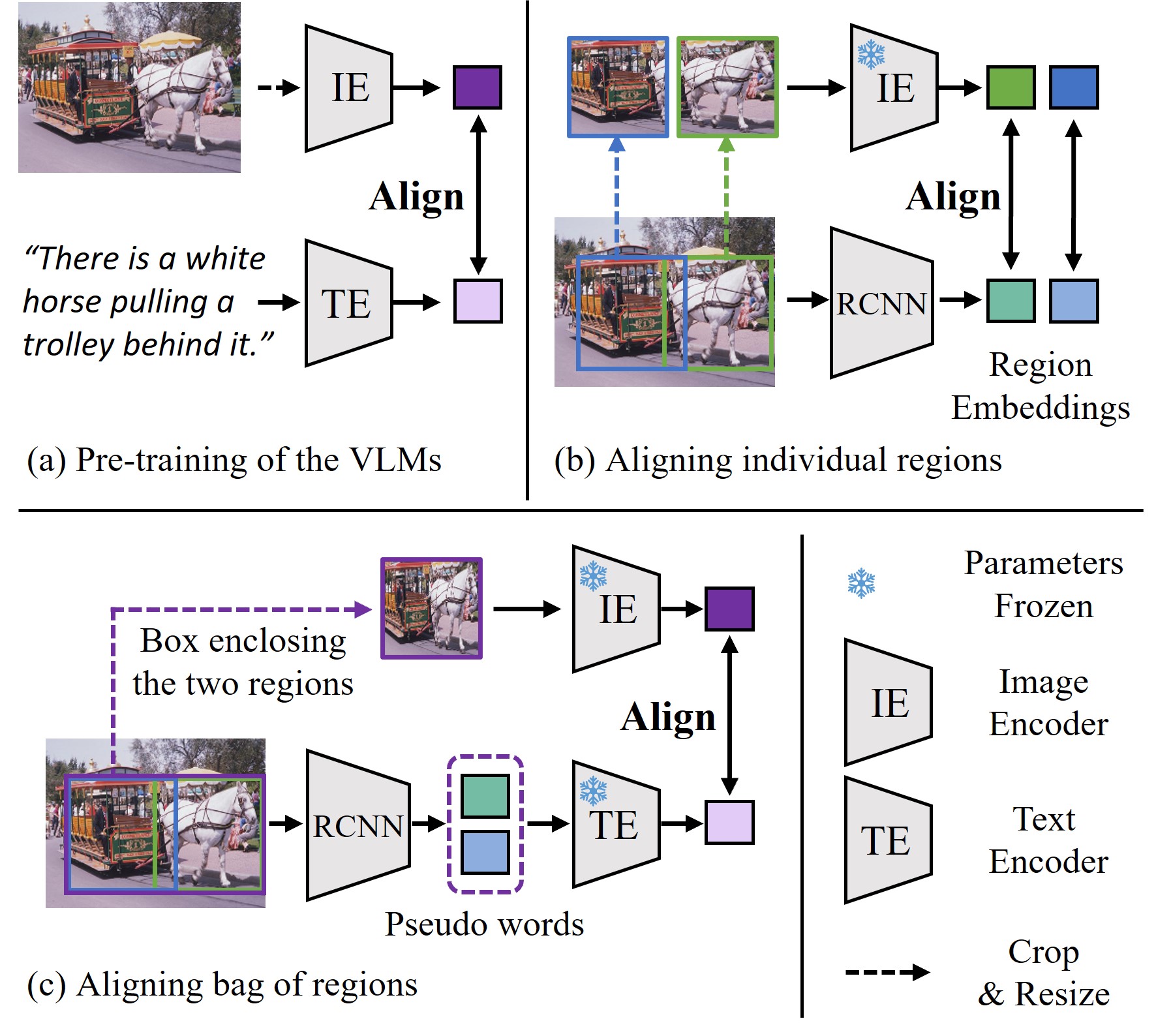}
	\caption{
	\textbf{(a)} Typical vision-language models (VLMs) learn to align representations of images and captions with rich compositional structure.
    \textbf{(b)} Existing distillation-based object detectors align each \emph{individual} region embedding to features extracted by the frozen image encoder of VLMs. 
    \textbf{(c)} Instead, the proposed method aligns the embedding of \emph{bag of regions}. The region embeddings in a bag are projected to the word embedding space (dubbed as pseudo words), formed as a sentence, and then sent to the text encoder to obtain the \emph{bag-of-regions} embedding, which is aligned to the corresponding image feature extracted by the frozen VLMs.
	}
	\vspace{-4mm}
	\label{fig:teaser}
\end{figure}

A traditional object detector can only recognize categories learned in the training phase, restricting its application in the real world with a nearly unbounded concept pool.
Open-vocabulary object detection (OVD), a task to detect objects whose categories are absent in training, has drawn increasing research attention in recent years.

A typical solution to OVD, known as the distillation-based approach, is to distill the knowledge of rich and unseen categories from pre-trained vision-language models (VLMs)~\cite{radford2021learning, jia2021scaling}.
In particular, VLMs learn aligned image and text representations on large-scale image-text pairs (Fig.~\ref{fig:teaser}(a)). Such general knowledge is beneficial for OVD. To extract the knowledge, most distillation-based approaches~\cite{gu2021open, Du_2022_CVPR, zang2022open} align each \emph{individual} region embedding to the corresponding features extracted from the VLM (Fig.~\ref{fig:teaser}(b)) with some carefully designed strategies.

We believe VLMs have implicitly learned the inherent compositional structure of multiple semantic concepts (\eg, co-existence of stuff and things~\cite{panoptic,Adelson2001}) from a colossal amount of image-text pairs. A recent study, MaskCLIP~\cite{zhou2022maskclip}, leverages such a notion for zero-shot segmentation.
Existing distillation-based OVD approaches, however, have yet to fully exploit the compositional structures encapsulated in VLMs.
Beyond the distillation of individual region embedding, we propose to align the embedding of \emph{BAg of RegiONs}, dubbed as \method. Explicitly learning the co-existence of visual concepts encourages the model to understand the scene beyond just recognizing isolated individual objects.

\method~is easy to implement.
As shown in Fig.~\ref{fig:teaser}(c), \method~first samples contextually interrelated regions to form a `bag'.
Since the region proposal network (RPN) is proven to cover potential novel objects~\cite{gu2021open, zhou2022detecting}, we explore a neighborhood sampling strategy that samples boxes around region proposals to help model the co-occurrence of a bag of visual concepts.
Second, \method~obtains the bag-of-regions embeddings by projecting the regional features into the word embedding space and encoding these pseudo words with the text encoder (TE) of a frozen VLM~\cite{radford2021learning}. By projecting region features to pseudo words, \method~naturally allows TE to effectively represent the co-occurring semantic concepts and understand the whole scene.
To retain the spatial information of the region boxes, \method~projects the box shape and box center position into embeddings and add to the pseudo words before feeding them to TE.

To train \method, the bag-of-regions embeddings are learned to be aligned to the embeddings obtained by feeding the image crops that enclose the bag of regions to the teacher, \ie, the image encoder (IE) of the VLM.
We adopt a contrastive learning approach~\cite{tian2020contrastive} to learn the pseudo words and the bag-of-regions embeddings.
Consistent with the VLMs' pre-training (\eg, CLIP~\cite{radford2021learning}), the contrastive loss pulls close corresponding student (the detector) and teacher (IE) embedding pairs and pushes away non-corresponding pairs.

We conduct extensive experiments on two challenging benchmarks, OV-COCO and OV-LVIS. The proposed method consistently outperforms existing state-of-the-art methods~\cite{zang2022open, Du_2022_CVPR, zhou2022detecting} in different settings. Combined with Faster R-CNN, {\method} achieves a 34.0 (4.6 increase) box AP$_\text{50}$ of novel categories on OV-COCO and 22.6 (2.8 increase) mask mAP of novel categories on OV-LVIS. It is noteworthy that \method~can also distill knowledge from caption supervision -- it achieves 32.7 box AP$_\text{50}$ of novel categories on OV-COCO, outperforming previous approaches that use COCO caption ~\cite{zareian2021open, zhong2022regionclip, zhou2022detecting, gao2021towards}.

\raggedbottom

%% file: 02_related.tex
\section{Related Work}
\label{sec:related}

\myparagraph{Vison-Language Pretraining and Its Applications.} Vision-language pre-training aims to learn aligned image and text representations~\cite{radford2021learning, jia2021scaling, frome2013devise, JayaramanG14, VILT} on large-scale image-text pairs.
There are many studies~\cite{VILT, BLIP, ALBEF, ViLBERT} that pre-train vision-language models (VLMs) to improve the performance of downstream recognition and generation tasks.
There are also studies that learn aligned vision-language representation so that the images can be classified with arbitrary texts~\cite{frome2013devise, ConvSE}.
Recent attempts~\cite{radford2021learning, jia2021scaling, LIT} push forward this direction by conducting contrastive learning in VLMs on billion-scale image-text pairs. These models show impressive zero-shot performance when they are transferred to image classification tasks.

Inspired by the success of VLMs~\cite{radford2021learning}, some works try to exploit the alignment of vision-language representations for dense prediction tasks, \eg segmentation~\cite{li2022language, zhou2022maskclip, rao2022denseclip} and detection~\cite{shi2022proposalclip, gu2021open, Du_2022_CVPR, GLIP}. In particular, MaskCLIP~\cite{zhou2022maskclip} shows that the image encoder in VLMs~\cite{radford2021learning} captures the stuff and things in a complex scene, where the pixel embeddings of each concept are naturally aligned with the corresponding text representations, although the original CLIP~\cite{radford2021learning} model does not explicitly learn this target.
This implies that VLMs, after trained on a massive amount of image-text pairs, have implicitly learned the compositional structure of multiple semantic concepts, which naturally exist in image-text pairs.
This motivates us to explore the representation alignment between bag of regions and bag of words, different from previous works~\cite{li2022language, gu2021open} that focus on aligning the representation of individual pixels, regions, or words in VLMs.

\myparagraph{Open-Vocabulary Object Detection.} 
Traditional object detectors~\cite{chen2019htc, ren2015faster, detr, zhu2020deformable, detection_survey} are limited to pre-defined object categories. To detect objects of unseen categories, zero-shot object detection (ZSD)~\cite{bansal2018zero, zheng2020zero,Rahman2019,synthesizing_zeroshot,DemirelCI18} is proposed to align individual region embeddings with the text embeddings of categories through different strategies.
Recent attempts further explore open-vocabulary object detection (OVD)~\cite{zareian2021open}, a more general form of ZSD that leverages weak supervisions like visual grounding data~\cite{GLIP}, image captions~\cite{zareian2021open, zhong2022regionclip, gao2021towards}, and image labels~\cite{zhou2022detecting}.
Large-scale pre-trained VLMs~\cite{radford2021learning} are also exploited for their remarkable zero-shot recognition ability.
These VLMs can generate conditional queries~\cite{zang2022open} or serve as a good teacher for knowledge distillation~\cite{gu2021open}.
Specifically, distillation-based approaches~\cite{gu2021open, Du_2022_CVPR} extract embeddings on pre-computed region proposals and \emph{individually} align them to the corresponding features obtained from the VLMs.
To our best knowledge, \method~ is the first attempt to lift the learning from \emph{individual} regions to the \emph{bag of regions} for OVD.

%% file: 03_method.tex
\begin{figure*}[t]
	\centering
	\includegraphics[width=1.0\textwidth]{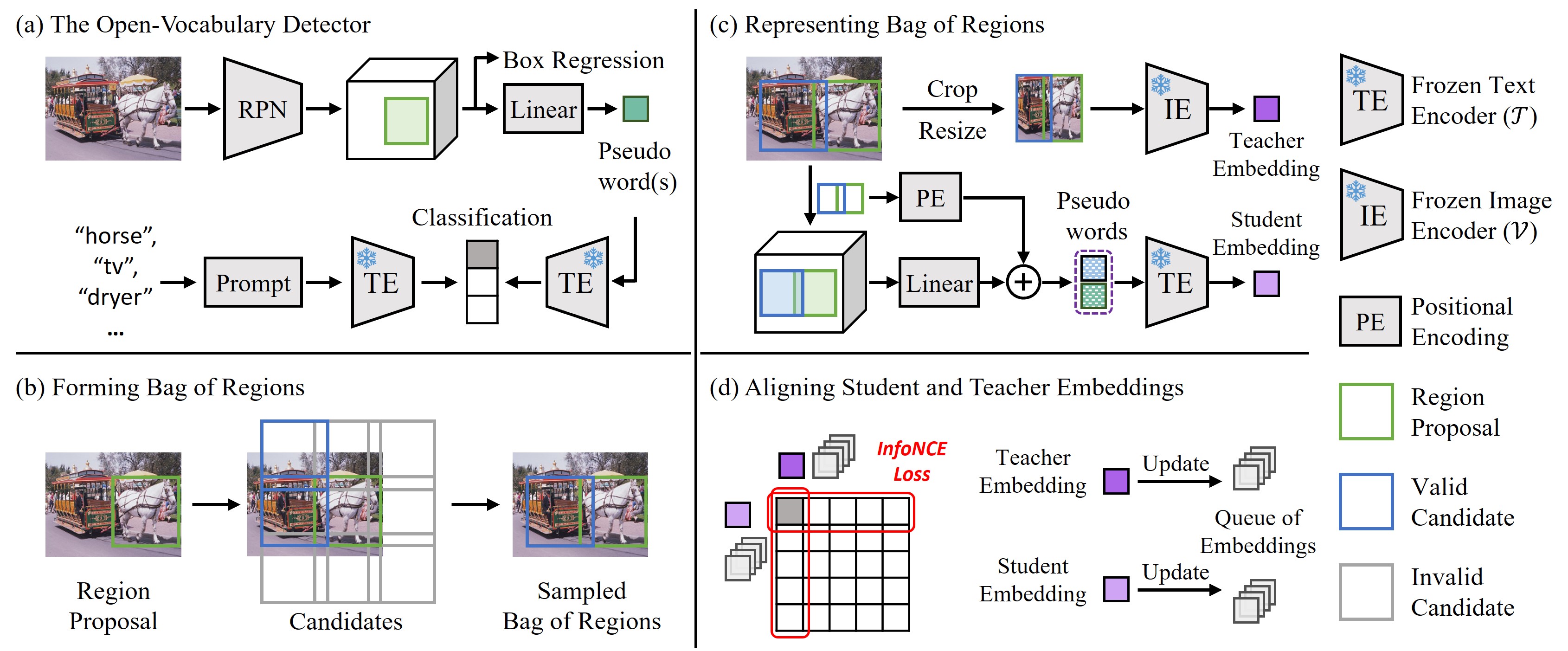}
	\caption{Overview of \method. \textbf{(a)} BARON is based on a Faster R-CNN whose classifier is replaced by a linear layer to map region features into pseudo words. 
     \textbf{(b)} BARON takes region proposals and its surrounding boxes to form bags of regions. 
     \textbf{(c)} BARON obtains student and teacher embeddings for the bag of regions from the pre-trained VLMs. \textbf{(d)} BARON learns the alignment by the InfoNCE loss and maintains queues of embeddings to provide sufficient negative pairs for InfoNCE loss.}
	\label{fig:overview}
\end{figure*}

\section{Method}
\label{sec:method}
Our method makes the first attempt to align embedding of \emph{bag of regions} beyond \emph{individual} regions for OVD. We call our method BARON.
In this work, we instantiate the idea of BARON based on the commonly used Faster-RCNN~\cite{ren2015faster} and modify it for OVD (Sec.~\ref{sec:baseline}).
We design a simple strategy to form the bag of regions from the region proposals (Sec.~\ref{sec:context_sampling}). 
The bag of regions is treated as a bag of words to obtain the bag-of-regions embedding (Sec.~\ref{sec:context_features}), which are then aligned with the corresponding features from VLMs (Sec.~\ref{sec:contrastive_loss}).
BARON is general that it can align the bag-of-regions embeddings to not only image representations but also text representations (Sec.~\ref{sec:caption_supervision}).

\subsection{Preliminaries}\label{sec:baseline}
In this paper, we instantiate the idea upon Faster R-CNN~\cite{ren2015faster} for simplicity.
The idea can also be used for other architectures~\cite{detr, lin2017focal} applicable for OVD.
To enable Faster R-CNN to detect objects from arbitrary vocabularies, we replace the original classifier with a linear layer that projects the region features into the word embedding space (dubbed as pseudo words) (Fig.~\ref{fig:overview}(a)). In practice, the linear layer maps a region feature to multiple pseudo words to represent the rich semantic information of each object, similar to those category names consisting of multiple words (\eg, horse-driven trolley).
Finally, we feed these pseudo words into the text encoder and then calculate the similarity with the category embeddings to obtain final classification results.

As shown in Fig.~\ref{fig:overview}(a), given $C$ object categories, we obtain the embedding $f_{c}$ for the $c$-th category by feeding the category names with a prompt template, \eg, \texttt{`a photo of \{category\} in the scene'} to the text encoder $\mathcal{T}$. For a region and its pseudo words $w$, the probability of the region to be classified as the $c$-th category is

\begin{equation}
\vspace{-5pt}
    p_c = \frac{\exp(\tau \cdot \langle \mathcal{T}(w), f_c \rangle)}{\sum_{i=0}^{C-1}\exp(\tau \cdot \langle \mathcal{T}(w), f_i\rangle)},
\label{eq:cls}
\end{equation}

\noindent where $\langle, \rangle$ denotes the cosine similarity and $\tau$ is the temperature to re-scale the value.

During training, only the boxes of base categories are annotated and the learning on base categories follows the convention of Faster R-CNN with regression and classification losses~\cite{ren2015faster}.
To learn to detect novel categories that do not have box annotations in training, previous distillation-based approaches~\cite{gu2021open, Du_2022_CVPR} \emph{individually} align region embeddings (\eg, $\mathcal{T}(w)$) to the corresponding features obtained from the VLMs. To further exploit the power of VLMs that capture the compositional structure of multiple concepts, we lift the learning from \emph{individual} regions to the \emph{bag of regions}.

\subsection{Forming Bag of Regions}
\label{sec:context_sampling}

Our framework is inspired by existing OVD approaches that distill knowledge from the image encoder of VLMs.
Specifically, we choose the image encoder of the VLMs as the teacher and expect it to teach the detector. But different from existing approaches, we wish the detector to learn the co-existence of multiple concepts, especially the potential existence of novel objects.
To effectively and efficiently learn such knowledge from the VLMs, we consider the following two properties of regions inside a bag:
1) the regions need to be close to each other because an image crop enclosing distanced regions will include a larger proportion of redundant image contents, distracting the image encoder from representing the bag of regions;
2) the regions should have similar sizes, as an imbalanced size ratio among regions will make the image representations dominated by the largest region.

\begin{figure}[ht]
    \centering
    \includegraphics[width=0.95\linewidth]{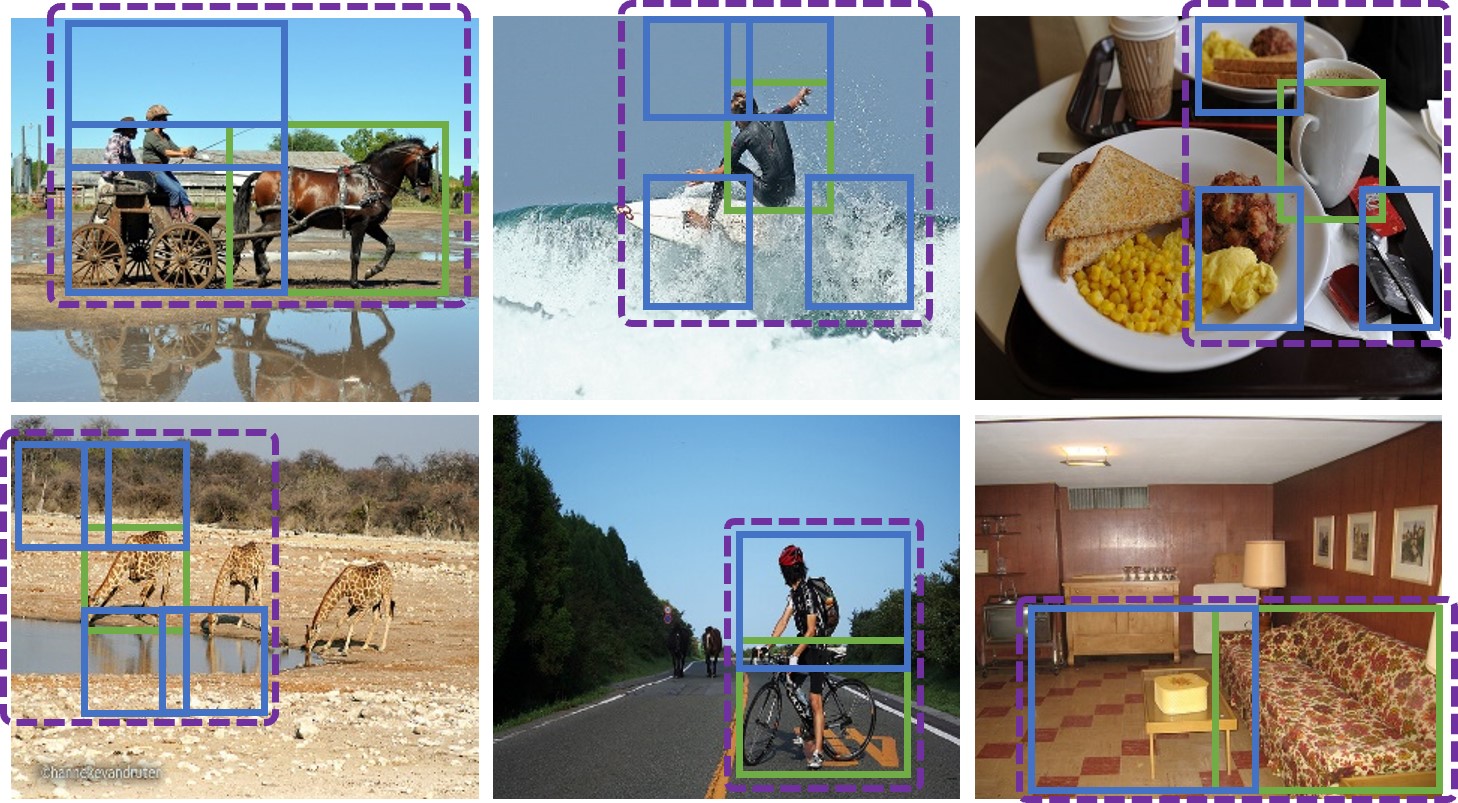}
    \vspace{-2pt}
    \caption{Visualization of sampled bags of regions. Green boxes denote the region proposals and blue boxes are sampled neighbors (candidates). Areas exceeding image boundary are cropped out.}
    \label{fig:sample}
    \vspace{-12pt}
\end{figure}

According to these two requirements, we adopt a simple neighborhood sampling strategy to form the bag of regions based on the region proposals predicted by the region proposal network (RPN)~\cite{ren2015faster}.
Concretely, for each region proposal, we take its surrounding eight boxes (neighbors) as the candidates, which are spatially close to each other as shown in Fig.~\ref{fig:overview}(b).
We also allow these candidates to overlap slightly at a specific Intersection over Foreground (IOF) to improve the continuity of regional representations.
To balance the sizes of regions inside the bag, we simply let the eight candidate boxes have the same shape as the region proposal.
In practice, candidates that exceed the boundary of the image over a certain proportion, \eg, more than $\frac{2}{3}$ of their area out of the image, will be discarded.
The remaining candidates are independently sampled, and the sampled boxes, together with the region proposal, form a bag of regions.

We sample $G$ groups for each region proposal to obtain rich bag-of-regions representations.
At each time of sampling, the probability of a candidate being sampled is adjusted to prevent the box enclosing the grouped regions from having an extreme aspect ratio. Assuming a base probability $p_b$ and a scaling factor $\alpha$, the probability to sample the left and right candidates is $p = p_b \times min\{(\frac{H}{W})^\alpha, 1\}$ and the probability to sample upper and bottom candidates is $p = p_b \times min\{(\frac{W}{H})^\alpha, 1\}$, where $H$ and $W$ are the height and width of the region proposal box. $\alpha$ is 3.0 by default. 

We show some sampling results of bags of regions on COCO dataset~\cite{lin2014microsoft} in Fig~\ref{fig:sample}. The grouped regions in a bag can cover objects (in blue) co-occurring with the object in the region proposal (in green). The context of the co-occurring objects leads BARON to get a scene-level understanding of the regions, \eg `a horse pulling the carriage' and `a man surfing on the wave'. Novel object categories occasionally appear in the bag of regions, \eg the `carriage' in the top left image of Fig~\ref{fig:sample} and the `cup' in the top right. In the following Sec~\ref{sec:context_features} and Sec~\ref{sec:contrastive_loss}, these potential novel object categories would be learned in the context of the bag of regions.

\subsection{Representing Bag of Regions}\label{sec:context_features}

With the sampled bags of regions, BARON obtains the bag-of-regions embeddings from both the student (\ie, the open-vocabulary object detector) and the teacher (\ie, VLMs).
We denote the $j$-th region in the $i$-th group as $b^i_j$ and the pseudo words after the projection layer as $w^i_j$.
For the pre-trained VLM, we use $\mathcal{T}$ to denote the text encoder and $\mathcal{V}$ to denote the image encoder.

\myparagraph{Student Bag-of-Regions Embedding.}
Because the region features are projected to word embedding space and learned to be aligned with the text embedding of category,
a straightforward way to obtain the embedding of a bag of regions is to concatenate the pseudo words and feed them to the text encoder $\mathcal{T}$.
However, the spatial information of the regions will be lost in such a process, including relative box center positions and relative box shapes.
The center position and shape indicate the spatial relationship among the regions in a bag, which is essential to induce a sentence-like interpretation for the bag of regions. And they are also encoded in the teacher (image encoder of VLMs) through positional embeddings in the input~\cite{radford2021learning}.
Therefore, BARON encodes the spatial information into positional embeddings $p^i_j$ that have the same dimension with $w^i_j$, following the practices of Transformers~\cite{ViT, transformer}.
The positional embeddings are added to the pseudo words before concatenation. Assuming the group size is $N^i$, this representation can be formulated as $f_t^i = \mathcal{T}(w^i_0+p^i_0, w^i_1+p^i_1, \dots, w^i_{N^i-1}+p^i_{N^i-1}).$ 

\myparagraph{Teacher Bag-of-Regions Embedding.} The image embedding of the grouped regions can be obtained by feeding the image crop that encloses the regions to the image encoder $\mathcal{V}$. The image crop may contain redundant contents that are outside the grouped regions; we mask out them in the attention layers of $\mathcal{V}$. The image feature can be formulated as $f_v^i = \mathcal{V}(b^i_0, b^i_1, \dots, b^i_{N_i-1})$.

\subsection{Aligning Bag of Regions}
\label{sec:contrastive_loss}
BARON aligns the bag-of-regions embeddings from student and teacher to make the student learn to encode the co-existence of multiple regions, which potentially contain multiple concepts.
We adopt the contrastive learning approach used in vision-language pre-training~\cite{radford2021learning}.
Specifically, given $G$ bags of regions, the alignment InfoNCE loss~\cite{oord2018representation} between bag-of-regions embeddings is calculated as

\begin{equation}
    \mathcal{L}_\mathrm{bag} = -\frac{1}{2} \sum_{k=0}^{G-1}(\log(p_{t, v}^k) + \log(p_{v, t}^k)).
\end{equation}
The $p_{t, v}^k$ and $p_{v, t}^k$ are calculated as 
\begin{align}
        p_{t, v}^k = \frac{\exp(\tau' \cdot \langle f_t^{k}, f_v^{k} \rangle)}{\sum_{l=0}^{G-1}\exp(\tau' \cdot \langle f_t^{k}, f_v^{l} \rangle)} \\
         p_{v, t}^k =  \frac{\exp(\tau' \cdot \langle f_v^{k}, f_t^{k} \rangle)}{\sum_{l=0}^{G-1}\exp(\tau' \cdot \langle f_v^{k}, f_t^{l} \rangle)},
\end{align}
respectively, where $\tau'$ is the temperature to re-scale the cosine similarity. The loss pulls positive pairs $\{f_t^{k}, f_v^{k}\}$ close to each other and pushes away negative pairs $\{f_t^{k}, f_v^{l}\} (k\neq l)$.

In practice, the number of groups $G$ is small for a single image. We maintain two queues to save the image and text embeddings in previous iterations during training to provide sufficient negative pairs~\cite{he2020momentum}.

\begin{table*}[ht]
\begin{center}
\caption{Comparison with state-of-the-art methods on OV-COCO benchmark. We separately compare our approach with methods distilling knowledge from CLIP and approaches using COCO caption. $\dagger$ means using proposals produced by MAVL~\cite{maaz2022class}.
}
\vspace{-6pt}
{\input{tables/coco_main}}
\label{tab:coco_main}
\end{center}
\vspace{-8pt}
\end{table*}

\myparagraph{Aligning Individual Regions.} 
The alignment between individual regions' student and teacher embeddings is complementary to that of a bag of regions. Therefore, we also adopt the individual-level distillation in our implementation. For computational efficiency, we obtain the teacher embeddings from the feature map of the image encoder's last attention layer by RoiAlign~\cite{he2017mask} instead of repeatedly passing image crops to the image encoder. Similarly, the student embeddings are extracted from the text encoder's last attention layer by averaging pseudo-word embeddings of the same region. We apply the InfoNCE loss and keep queues of embeddings to calculate the individual-level loss $\mathcal{L}_\mathrm{individual}$.

\begin{figure}[t]
    \begin{center}
        \vspace{-2pt}
    \includegraphics[width=0.75\linewidth]{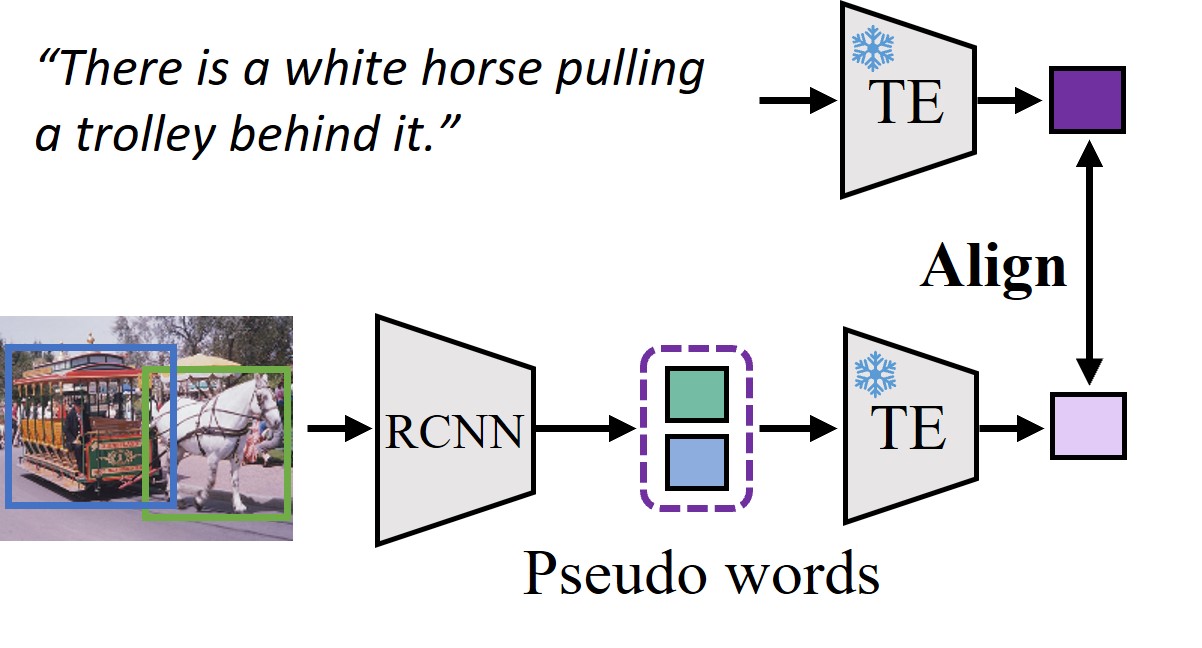}
    \vspace{-17pt}
    \end{center}
    \caption{The caption-version \method. We align the text embeddings of the bag of regions to the caption embeddings.} 
    \label{fig:caption}
    \vspace{-10pt}
\end{figure}

\subsection{Caption Supervision}\label{sec:caption_supervision}
It is noteworthy that BARON can be applied to caption supervision. As shown in Fig.~\ref{fig:caption}, the core idea is to replace the image embedding $f_v$ with embedding obtained by feeding the image captions to the text encoder $\mathcal{T}$.
To obtain the region groups, we randomly sample some region proposals generated by the RPN. As an image can have multiple captions, we follow the practice in UniCL~\cite{yang2022unified} to apply a soft cross entropy loss that simultaneously aligns the student bag-of-regions embedding to multiple caption embeddings. 
The alignment of individual regions is discarded since we cannot get the correspondence between pseudo words of regions and the actual words in a caption without grounding.
In this way, BARON now learn to align the bag-of-regions embedding of the image caption, which also describes the existence of multiple concepts of the image.

%% file: tables/coco_main.tex
 \tablestyle{3pt}{1.1}
  \scalebox{0.9}{
\begin{tabular}{l |c |c| c|c >{\color{gray}}c>{\color{gray}} c}
\toprule
Method &Supervision & Backbone & Detector & AP$_{50}^{\text{novel}}$ & AP$_{50}^{\text{base}}$ & AP$_{50}$ \\\shline
ViLD~\cite{gu2021open} & CLIP &  ResNet50-FPN  & FasterRCNN & 27.6& 59.5& 51.2\\
OV-DETR~\cite{zang2022open} & CLIP & ResNet50 & DeformableDETR & 29.4& 61.0& 52.7 \\
BARON (Ours) & CLIP & ResNet50-FPN  & FasterRCNN & \textbf{34.0}& 60.4& 53.5 \\ \hline
OVR-CNN~\cite{zareian2021open} & Caption & ResNet50-C4  & FasterRCNN & 22.8&46.0 &39.9 \\
RegionCLIP~\cite{zhong2022regionclip}  & Caption & ResNet50-C4  & FasterRCNN &26.8 & 54.8&47.5 \\
Detic~\cite{zhou2022detecting} & Caption & ResNet50-C4  & FasterRCNN & 27.8&51.1 &45.0 \\
PB-OVD~\cite{gao2021towards} & Caption & ResNet50-C4  & FasterRCNN &30.8 &46.1 &42.1 \\
VLDet~\cite{VLDet} & Caption & ResNet50-C4 & FasterRCNN & 32.0 & 50.6 & 45.8 \\
BARON (Ours) & Caption & ResNet50-C4 & FasterRCNN &\textbf{33.1} &54.8 &49.1 \\
\hline
Rasheed \etal~\cite{rasheedbridging}$^\dagger$ & CLIP + Caption & ResNet50-C4  & FasterRCNN & 36.6 & 54.0 & 49.4 \\
BARON (Ours)$^\dagger$ & CLIP + Caption & ResNet50-C4  & FasterRCNN &\textbf{42.7} & 54.9 & 51.7 \\
\bottomrule
\end{tabular}}

%% file: 04_experiments.tex
\section{Experiments}
\label{sec:exp}

\myparagraph{Datasets.} We evaluate our method on the two popular object detection datasets, \ie, COCO~\cite{lin2014microsoft} and LVIS~\cite{gupta2019lvis}. For the COCO dataset, we follow OV-RCNN~\cite{zareian2021open} to split the object categories to 48 base categories and 17 novel categories. For the LVIS dataset, we follow ViLD~\cite{gu2021open} to split the 337 rare categories into novel categories and the rest common and frequent categories into base categories. For brevity, we denote the open-vocabulary benchmarks based on COCO and LVIS as OV-COCO and OV-LVIS. 

\myparagraph{Evaluation Metrics.} We evaluate the detection performance on both base and novel categories for completeness. 
For OV-COCO, we follow OV-RCNN~\cite{zareian2021open} to report the box AP at IoU threshold 0.5, noted as AP$_{50}$. 
For OV-LVIS, we report both the mask and box AP averaged on IoUs from 0.5 to 0.95, noted as mAP.
The AP$_{50}$ of novel categories (AP$_{50}^{\mathrm{novel}}$) and mAP of rare categories (AP$_{r}$) are the main metrics that evaluate the open-vocabulary detection performance on OV-COCO and OV-LVIS, respectively. 

\myparagraph{Implementation Details.} 
We build \method~on Faster R-CNN~\cite{ren2015faster} with ResNet50-FPN~\cite{lin2017fpn}.
For a fair comparison with existing methods, we initialize the backbone network with weights pre-trained by SOCO~\cite{wei2021aligning} and apply synchronized Batch Normalization (SyncBN)~\cite{megdet} following DetPro~\cite{Du_2022_CVPR}. We choose the $2 \times$ schedule (180, 000 iterations) for the main experiments on COCO and LVIS~\cite{mmdetection, wu2019detectron2}. For the pre-trained VLM, we choose the CLIP~\cite{radford2021learning} model based on ViT-B/32~\cite{ViT}. For the adaptation to caption supervision, we base our method on Faster R-CNN with ResNet50-C4 backbone~\cite{ren2015faster} and adopt the $1\times$ schedule. For the prompt of category names, we use the hand-crafted prompts in ViLD~\cite{gu2021open} for all our experiments by default.
We use learned prompt only when comparing with DetPro~\cite{Du_2022_CVPR}.

\begin{table*}[t]
  \small
  \centering
\caption{Comparison with state-of-the-art methods on OV-LVIS. * denotes the re-implemented ViLD~\cite{gu2021open} reported in DetPro~\cite{Du_2022_CVPR}.
}
\vspace{-8pt}
 {\input{tables/lvis_main}}
\label{tab:lvis_main}
\end{table*}

\begin{table*}[t]
  \small
  \centering
\vspace{-2pt}
\caption{Comparison of the transfer ability of the model trained on OV-LVIS. * denotes the re-implemented ViLD~\cite{gu2021open} reported in DetPro~\cite{Du_2022_CVPR}. $\ddagger$ denotes that we use hand-crafted prompts for a fair comparison with ViLD.}
\vspace{-8pt}
  {\input{tables/transfer}}
\vspace{-10pt}
\label{tab:transfer}
\end{table*}

\subsection{Benchmark Results}
\myparagraph{OV-COCO.}
We report the comparison with previous methods in Table~\ref{tab:coco_main}.
BARON surpasses previous state of the arts with either pre-trained VLMs or COCO captions~\cite{chen2015microsoft}, indicating its effectiveness and flexibility.
Note that OV-DETR is based on the Deformable DETR~\cite{zhu2020deformable}, which is stronger than the Faster R-CNN~\cite{ren2015faster} with higher performance on base categories. 
But BARON still outperforms OV-DETR by 4.6 AP$_{50}$ on novel categories.
When using caption supervision, BARON even outperforms PB-OVD~\cite{gao2021towards} that uses sophisticated pseudo-labeling. Combining CLIP image features, COCO captions, and MAVL~\cite{maaz2022class} proposals, BARON significantly outperforms Rasheed \etal~\cite{rasheedbridging} by a large margin.

\myparagraph{OV-LVIS.}
We compare BARON with other methods on the OV-LVIS benchmark in Table~\ref{tab:lvis_main}.
Because ViLD~\cite{gu2021open} is trained with large-scale jittering~\cite{simple_copy_paste} and a prohibitive $32\times$ schedule, DetPro~\cite{Du_2022_CVPR} re-implemented it with backbone weights pre-trained by SOCO~\cite{wei2021aligning} and a regular $2\times$ schedule.
DetPro also proposes learned prompts for the category's names. Besides, an ensembling strategy for classification scores is adopted in ViLD and DetPro. 
For fair comparison, we respectively implement our method on OV-LVIS with and without the these tricks. 
\method~achieves the best performance in all the scenarios and can even surpass ViLD that adopts the ensembling strategy without these tricks.

\myparagraph{Transfer to Other Datasets.}
We transfer the open-vocabulary detector trained on OV-LVIS to three other datasets, including Pascal VOC 2007~\cite{everingham2010pascal} test set, COCO~\cite{lin2014microsoft} validation set and Objects365~\cite{shao2019objects365} v2 validation set. We compare our results with DetPro~\cite{Du_2022_CVPR} and the ViLD~\cite{gu2021open} implemented in DetPro. The comparison is fair since all the models are based on the same object detector and training schedule. 
As shown in Table~\ref{tab:transfer}, our approach exhibits better generalization ability on all of the three datasets.

\subsection{Ablation Study}
In this section, we ablate the effectiveness of components in \method on OV-COCO benchmark.

\begin{table*}[h]
\centering
\hspace{1mm}
\begin{minipage}[t]{0.31\textwidth}
\centering
\caption{Effectiveness of main components of \method}
\vspace{-6pt}
\scalebox{0.9}{\input{tables/ablation_overall}}
\label{tab:ablation_overall}
\end{minipage}
\hspace{6mm}
\begin{minipage}[t]{0.31\textwidth}
\centering
\caption{Exploring sampling strategies to obtain bag of regions}
\vspace{-6pt}
\scalebox{0.9}{\input{tables/ablation_sampling}}
\label{tab:sampling_baseline}
\end{minipage}
\hspace{1mm}
\begin{minipage}[t]{0.31\textwidth}
\centering
\caption{Overlap (IOF) between regions}
\vspace{-6pt}
\scalebox{0.916}{\input{tables/ablation_4}}
\label{tab:overlap}
\end{minipage}
\end{table*}

\begin{table*}[h]
\vspace{-4pt}
\hspace{1mm}
\centering
\begin{minipage}[t]{0.31\textwidth}
\centering
\caption{Ablation study on the sampling probability $p_b$}
\vspace{-8pt}
\scalebox{0.95}{\input{tables/ablation_2}}
\label{tab:probability}
\end{minipage}
\hspace{6mm}
\begin{minipage}[t]{0.31\textwidth}
\centering
\caption{Number of sampled bags per region proposal}
\vspace{-8pt}
\scalebox{0.95}{\input{tables/ablation_3}}
\label{tab:groups}
\end{minipage}
\hspace{1mm}
\begin{minipage}[t]{0.31\textwidth}
\centering
\caption{Number of pseudo words}
\vspace{-8pt}
\scalebox{0.96}{\input{tables/ablation_5}}
\label{tab:word_num}
\end{minipage}
\vspace{-7pt}
\end{table*}

\myparagraph{Effectiveness of Aligning Bag of Regions.} We start from a baseline that only uses the individual-level loss $\mathcal{L}_{\text{individual}}$. As shown in Table~\ref{tab:ablation_overall}($\#1$), the individual-level baseline achieves 25.7 mAP$_{\text{50}}$ on novel categories. We then replace $\mathcal{L}_{\text{individual}}$  with the loss for bag of regions $\mathcal{L}_{\text{bag}}$ (Table~\ref{tab:ablation_overall}($\#2$)). Without considering the spatial information of region boxes, the performance on novel categories is compatible with aligning individual regions.
When adding the positional embedding of the region boxes' spatial information, the performance on novel categories (Table~\ref{tab:ablation_overall}($\#3$)) dramatically increases by 7.1 mAP$_{\text{50}}$.
This means that the spatial information is essential to effectively exploit the compositional structure of co-occurring visual concepts in a bag of regions.
Finally, in Table~\ref{tab:ablation_overall}($\#4$), we find that the individual-level loss is complementary to the bag-of-regions alignment, which brings 1.2 mAP$_{\text{50}}$ performance gain on novel categories.

\myparagraph{Sampling Strategies.}
We explore two baselines to sample bags of regions to support the rationale of our neighborhood sampling strategy. The first is to equally split an image into grids (dubbed as grid sampling) like the pre-training stage in OVR-CNN~\cite{zareian2021open} such that the fixed grids form a bag of regions. And the second is to
randomly sample region proposals to form a bag of regions (dubbed as random sampling). For a fair comparison, we keep the number of sampled regions in each sampling strategy roughly the same. Concretely, we split the images to $3 \times 3 = 9$ grids for the grid sampling strategy and sample 9 region proposals for the random sampling strategy. For these two strategies, we take 4 permutations of the regions to obtain richer bag-of-regions embeddings so that there would be 36 regions for each image. For the neighborhood sampling, we introduce a reduced version of the strategy that restricts the number of region proposals per image to 12 and takes 1 bag per proposal. This is because we record that the average number of regions in a bag is 3 with $p_b=0.3$ and $\alpha=3.0$, meaning the average sampled regions for each image in the neighborhood sampling strategy is close to $3 \times 12 = 36$. 

The grid sampling strategy, whose fixed regions may either contain too many objects or only small parts of an object, achieves 25.4 mAP$_{50}$ on the novel categories as shown in Table~\ref{tab:sampling_baseline}($\#1$). For the random sampling strategy, regions in a bag have different sizes and shapes and the distance between the regions can be large, which hinders the image encoder to exactly represent the bag of regions. It achieves 27.3 mAP$_{50}$ as shown in Table~\ref{tab:sampling_baseline}($\#2$). While our neighborhood sampling strategy utilizes the same amount of regions, we record 32.2 mAP on the novel categories in Table~\ref{tab:sampling_baseline}($\#3$). Compared to these two baselines, our neighborhood sampling strategy captures potential objects in the vicinity of region proposals and ensures the teacher embedding exactly represents the bag of regions by sampling the neighboring boxes of region proposals. Note that our result in Table~\ref{tab:sampling_baseline}($\#4$) is achieved with 3 groups per region proposal and no limitation on the number of region proposals. More details on how we develop the sampling strategy are in the appendix.

\myparagraph{Box Overlap between Regions in a Bag.} We let the sampled regions in a group overlap at a certain IOF. In Table~\ref{tab:overlap}, we show how the overlap between boxes affects the performance. The scalar smaller than 0.0 in Table~\ref{tab:overlap}($\#1$) means that we keep an interval between the sampled boxes. The best performance (34.0 mAP$_{\text{50}}$) comes with an overlap of 0.1. Table~\ref{tab:overlap}($\#1-3$) indicate that the regions in a group need to have a continuity of semantics. Table~\ref{tab:overlap}($\#3-5$) indicate that the regions also need to cover diverse image contents.

\myparagraph{Sampling Probability.} We study the effect of the probability $p_b$ to sample the candidate boxes, which affects the number of regions in a bag. The details are in Sec.~\ref{sec:context_sampling}. Note that the $p_b$ will be adjusted by the aspect ratio scaled with the scale factor $\alpha$. We fix $\alpha$ as 3.0 and sample 3 bags for each region proposal. We observe that the best result on novel categories (34.0 mAP$_{50}$) is achieved with $p_b=0.3$ in Table~\ref{tab:probability}.

\myparagraph{Number of Bags Per Proposal.} We show the effect of the number of sampled bags ($\#$bags) for each region proposal in Table~\ref{tab:groups}. For details, please refer to Sec~\ref{sec:context_sampling}. We fix $\alpha$ as 3.0 and the sampling probability $p_b$ as 0.3. We observe that the best result on novel categories (34.0 mAP$_{50}$) is achieved with three bags of regions for each region proposal in Table~\ref{tab:groups}.

\myparagraph{Number of Pseudo Words Per Region.} As an object category often needs many words to reach a precise description, we study the number of pseudo words ($\#$wordss) predicted for each region of interest. Table~\ref{tab:word_num}($\#1-3$) show that the performance on novel categories increases with the number of pseudo words, indicating that stacking more pseudo words to a certain extent can strengthen the detector's ability to distinguish object categories. However, the results in Table~\ref{tab:word_num}($\#1-3$) show that further increasing the number of pseudo words does not bring performance gain, and redundant words can even do harm to the performance.

\begin{figure*}[ht]
    \centering
    \includegraphics[width=1.0\linewidth]{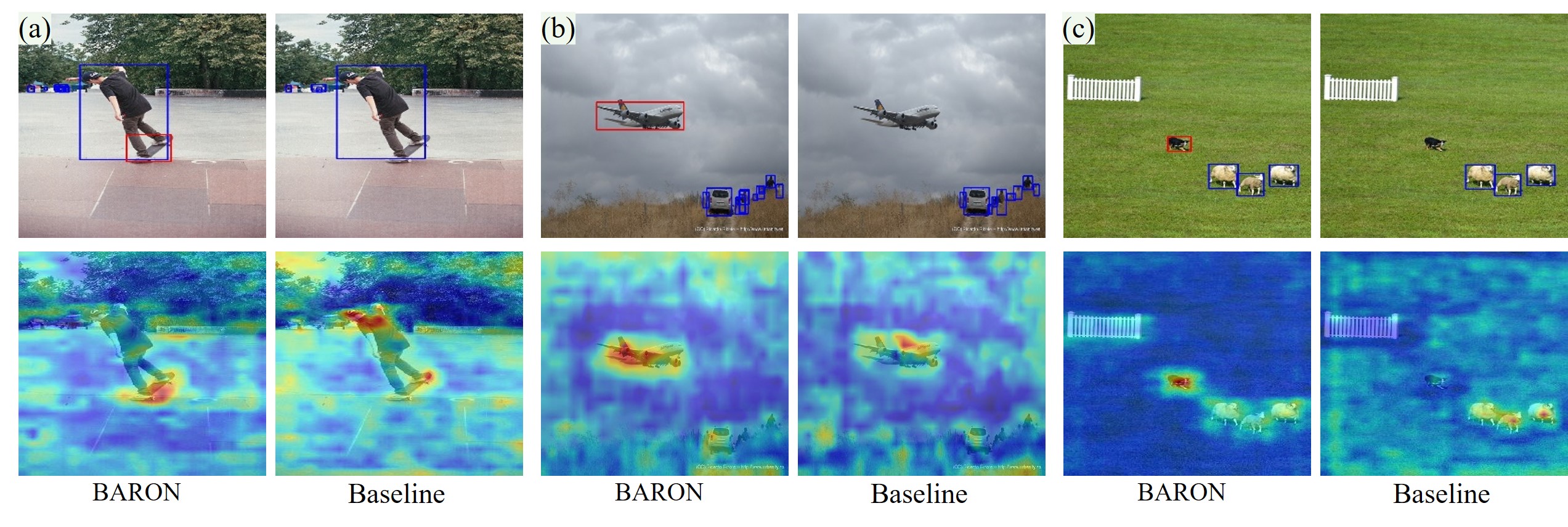}
    \vspace{-24pt}
    \caption{Qualitative comparisons of BARON and the individual-level baseline in Table~\ref{tab:ablation_overall}($\#1$).
    \textbf{Top:} Red boxes are for the novel categories and blue for the base categories. \textbf{Bottom:} Feature map's responses to the queried novel object categories. From (a) to (c), the queried novel categories are `skateboard', `airplane' and `dog', respectively. 
    \method~detects objects of novel categories that are missed by the baseline. 
    }
    \label{fig:vis}
    \vspace{-13pt}
\end{figure*}

\begin{figure}[h]
    \centering
    \includegraphics[width=1.0\linewidth]{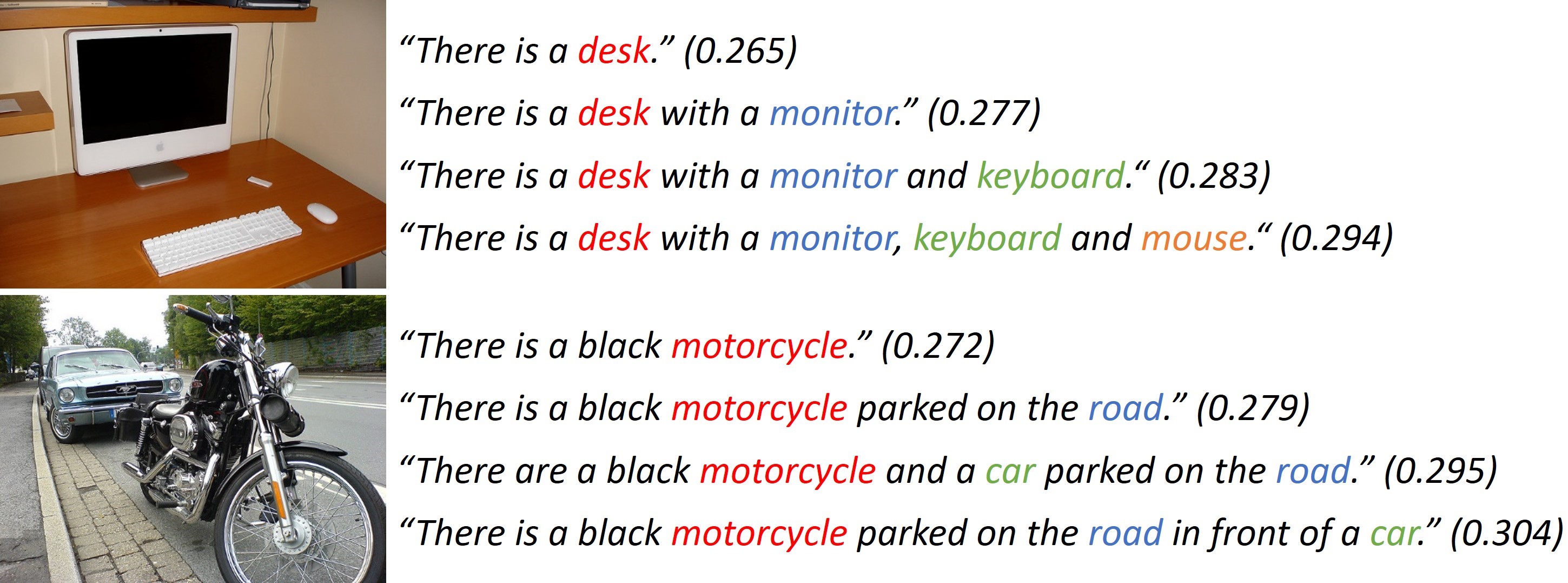}
    \vspace{-12pt}
    \caption{For each image, we incrementally add the object categories that appear in it to the text description. The similarity score between the image and text embeddings increases as the text description becomes more complete and precise.}
    \label{fig:motivation}
 \vspace{-7pt}
\end{figure}

\subsection{Further Analysis}
We show qualitative results in this section to further analyze the effectiveness of our method.

\myparagraph{Co-occurrence of Objects.} We first illustrate how a pre-trained VLM~\cite{radford2021learning} captures the co-occurrence of objects by comparing the image-text similarity in Fig.~\ref{fig:motivation}. 
We use the CLIP-ViT-B/32 model pre-trained on more than 400 million image-text pairs to obtain the image and text embeddings.
For each image, we incrementally add the object categories that appear to the text description. 
We observe that the similarity score between the image and text embeddings increases as the text description includes more concepts as shown in Fig.~\ref{fig:motivation}.
The examples reveal that the large-scale VLMs could capture the co-occurrence of multiple concepts in an image, although they are not explicitly trained to do so.
We believe this is because each image-text pair naturally contains multiple concepts and the VLMs could implicitly learn the underlying connections when training on massive-scale image-text pairs. We also find simple relationship between objects can also be captured by the VLMs, \eg, the similarity increases when we add the relation `in front of' to the text description in the second example of Fig~\ref{fig:motivation}.

\myparagraph{Visualization.} We further visualize the predictions of detectors learned through \method~and the individual-level baseline in Fig.~\ref{fig:vis}. 
The images are from COCO's validation set. We visualize the feature map's response to the novel categories using the Grad-CAM++~\cite{chattopadhay2018grad}.
We find that the model learned through \method~generates responses at locations of novel categories while the individual-level baseline induces weaker, incomplete or diffused responses.
We also notice that semantically related objects can respond to the queried object category.
For example, in Fig~\ref{fig:vis}(c) the location of a flock of sheep chased by a dog responds to the queried object `dog'. Note the `dog' and the `flock of sheep' are not in neighboring regions and the `dog' is much smaller in size. Even though we form bags of regions by neighboring regions of equal size in training, the model has the ability to capture the relationship between objects with imbalanced sizes or large distance. This phenomenon resembles the generalization ability of human language, where learned concepts can be applied to describe or recognize new things.

%% file: tables/lvis_main.tex
	
 \tablestyle{3pt}{1.1}
  \scalebox{0.9}{
\begin{tabular}{l|c|c|c>{\color{gray}}c>{\color{gray}}c>{\color{gray}}c|c>{\color{gray}}c>{\color{gray}}c>{\color{gray}}c}
  \toprule
 \multirow{2}{*}{Method}&\multirow{2}{*}{Ensemble}&\multirow{2}{*}{Learned Prompt} &\multicolumn{4}{c|}{Object Detection}&\multicolumn{4}{c}{Instance segmentation}\\
  &&&AP$_{r}$&AP$_{c}$&AP$_{f}$&AP&AP$_{r}$&AP$_{c}$&AP$_{f}$&AP\\ \shline
ViLD~\cite{gu2021open}& - & - & 16.3 &21.2 &31.6& 24.4 &16.1 &20.0 &28.3 &22.5\\
OV-DETR~\cite{zang2022open}& - &- & - & - & - & - & 17.4 &25.0& 32.5&26.6 \\
BARON (Ours)& - & - & \textbf{17.3} &25.6 &31.0 &26.3 &\textbf{18.0}&24.4 &28.9 &25.1 \\ \hline
ViLD~\cite{gu2021open} & \checkmark &- &16.7&26.5&34.2&27.8 & 16.6 & 24.6 & 30.3 & 25.5\\
ViLD*~\cite{gu2021open} & \checkmark &- &17.4&27.5&31.9&27.5&16.8&25.6&28.5&25.2\\
BARON (Ours) &\checkmark & -& \textbf{20.1} &28.4 &32.2 &28.4 &\textbf{19.2} &26.8 &29.4 &26.5\\\hline
DetPro~\cite{Du_2022_CVPR} &\checkmark & \checkmark& 20.8 &27.8&32.4&28.4&19.8 &25.6&28.9&25.9\\
BARON (Ours) & \checkmark & \checkmark &\textbf{23.2}&29.3&32.5&29.5&\textbf{22.6}&27.6&29.8&27.6\\
\bottomrule
\end{tabular}
}

%% file: tables/transfer.tex
 \tablestyle{3pt}{1.1}
  \scalebox{0.9}{
  \begin{tabular}{l|cc|cccccc|cccccc}
  \toprule
 \multirow{2}{*}{Method}&\multicolumn{2}{c|}{Pascal VOC}&\multicolumn{6}{c|}{COCO}&\multicolumn{6}{c}{Objects365}\\
  &AP$_{50}$&AP$_{75}$&AP&AP$_{50}$&AP$_{75}$&AP$_s$&AP$_m$&AP$_l$&AP&AP$_{50}$&AP$_{75}$&AP$_s$&AP$_m$&AP$_l$\\
  \midrule
Supervised~\cite{Du_2022_CVPR}&78.5&49.0&46.5&67.6&50.9&27.1&67.6&77.7&25.6&38.6&28.0&16.0&28.1&36.7\\
  \midrule
ViLD*~\cite{gu2021open}&73.9&57.9&34.1&52.3&36.5&21.6&38.9&46.1&11.5&17.8&12.3&4.2&11.1&17.8\\
BARON (Ours)$^\ddagger$&\textbf{74.5}&\textbf{57.9}&\textbf{36.3}&\textbf{56.1}&\textbf{39.3}&\textbf{25.4}&\textbf{39.5}&\textbf{48.2}&\textbf{13.2}&\textbf{20.0}&\textbf{14.0}&\textbf{4.8}&\textbf{12.7}&\textbf{20.1}\\
 \midrule
DetPro~\cite{Du_2022_CVPR} &74.6&57.9&34.9&53.8&37.4&22.5&39.6&46.3&12.1&18.8&12.9&4.5&11.5&18.6\\
BARON (Ours) &\textbf{76.0}&\textbf{58.2}&\textbf{36.2}&\textbf{55.7}&\textbf{39.1}&\textbf{24.8}&\textbf{40.2}&\textbf{47.3}&\textbf{13.6}&\textbf{21.0}&\textbf{14.5}&\textbf{5.0}&\textbf{13.1}&\textbf{20.7}\\

\bottomrule
\end{tabular}
}

%% file: tables/ablation_overall.tex
 \tablestyle{3pt}{1.1}
  \scalebox{0.9}{

\begin{tabular}{c |c |c| c|c c c}
 \midrule
\#  & $\mathcal{L}_\mathrm{individual}$& $\mathcal{L}_\mathrm{bag}$& PE & AP$_{50}^{\text{novel}}$ & AP$_{50}^{\text{base}}$ & AP$_{50}$ \\
 \midrule
1 & \checkmark & - & - & 25.7&59.6 &50.6 \\
2 & -  & \checkmark & - &25.7& 59.4&50.5 \\
3 & - & \checkmark  & \checkmark &32.8 & 60.1 &53.0\\
4 & \checkmark & \checkmark & \checkmark &\textbf{34.0} &60.4 &53.5 \\

\hline
\end{tabular}
}

%% file: tables/ablation_sampling.tex
 \tablestyle{3pt}{1.1}
  \scalebox{0.9}{\begin{tabular}{c|c| cc c}
 \midrule
$\#$&Sampling strategy & AP$_{50}^{\text{novel}}$ & AP$_{50}^{\text{base}}$ & AP$_{50}$\\
 \midrule
1&Grid & 25.4&58.0 & 49.5 \\
2&Random & 27.3 & 53.3 & 46.5 \\
3&Ours (reduced) & \textbf{32.2} &58.3 &51.5\\
4&Ours & \textbf{34.0} &60.4 &53.5\\
\hline
\end{tabular}
}

%% file: tables/ablation_4.tex
 \tablestyle{3pt}{1.1}
  \scalebox{0.9}{
  \begin{tabular}{c|c| cc c}
 \midrule
\#&Overlap & AP$_{50}^{\text{novel}}$ & AP$_{50}^{\text{base}}$ & AP$_{50}$\\
 \midrule
1 & -0.1& 32.5 & 59.7 & 52.6\\
2 & 0.0 & 33.6 & 60.2 & 53.2\\
3 & 0.1 & \textbf{34.0} & 60.4 & 53.5\\
4 & 0.2 & 33.8 & 59.8 & 53.0\\
5 & 0.3 & 33.7 & 60.0 & 53.1\\
\hline
\end{tabular}
}

%% file: tables/ablation_2.tex
 \tablestyle{3pt}{1.1}
  \scalebox{0.9}{
\begin{tabular}{c| cc c}
 \midrule
p$_b$ & AP$_{50}^{\text{novel}}$ & AP$_{50}^{\text{base}}$ & AP$_{50}$\\
 \midrule
0.1 & 33.7&59.8 &52.9\\
0.3 & \textbf{34.0} &60.4 &53.5\\
0.5 & 33.2 &60.0 &53.0\\
\hline
\end{tabular}
}

%% file: tables/ablation_3.tex
 \tablestyle{3pt}{1.1}
  \scalebox{0.9}{
\begin{tabular}{c| cc c}
 \midrule
\#\text{bags} & AP$_{50}^{\text{novel}}$ & AP$_{50}^{\text{base}}$ & AP$_{50}$\\
 \midrule
1 & 32.6 &58.5 &51.7\\
3 & \textbf{34.0} &60.4 &53.5\\
5 & 33.2 &60.0 &53.0\\
\hline
\end{tabular}
}

%% file: tables/ablation_5.tex
 \tablestyle{3pt}{1.1}
  \scalebox{0.9}{
  \begin{tabular}{c|c| cc c}
 \midrule
\#& \#words & AP$_{50}^{\text{novel}}$ & AP$_{50}^{\text{base}}$ & AP$_{50}$\\
 \midrule
1 &2 & 31.6 & 59.5 & 52.2 \\
2 &4 & 33.1 & 60.1 & 53.0 \\
3 &6 & \textbf{34.0} & 60.4 & 53.5 \\
4 &8 & 33.5 & 59.9 & 53.0 \\
\hline
\end{tabular}
}

%% file: 10_conclusion.tex
\section{Discussion and Conclusion}
\label{sec:conclusion}
This paper goes beyond the learning of individual regions to bag of regions in OVD, exploring the ability of large-scale VLMs to represent the compositional structure of multiple concepts that naturally exists in image-text pairs.
We develop a neighborhood sampling strategy to group contextually related regions into a bag and adopt the contrastive learning approach to align the bag-of-regions representations of the detector and pre-trained VLMs, which achieves new state-of-the-art performance on multiple OVD benchmarks.

The compositional structure explored in this paper is mainly about the co-occurrence of objects, and behaves like bag of words~\cite{mert2022when, ajinkya2021fistful}.
The more complex compositional structure in the language is still under-explored and whether modern pre-trained vision-language models capture such structure still remains an open problem for the community.
We look forward to further unveiling the behavior of the VLMs, and more importantly endowing the VLMs with human-like compositional representation to move to more generalized intelligence.

\textbf{Acknowledgement.}
This study is supported under the RIE2020 Industry Alignment Fund Industry Collaboration Projects (IAF-ICP) Funding Initiative, as well as cash and in-kind contribution from the industry partner(s). The work is also suported by Singapore MOE AcRF Tier 2 (MOE-T2EP20120-0001) and NTU NAP Grant. We also thank Lumin Xu and Yuanhan Zhang for valuable discussions.

\clearpage

%% file: 12_appendix.tex
\appendix
\label{sec:appendix}

\setcounter{table}{0}
\renewcommand{\thetable}{A\arabic{table}}
\setcounter{figure}{0}
\renewcommand{\thefigure}{A\arabic{figure}}
\setcounter{section}{0}
\renewcommand{\thesection}{A\arabic{section}}
\input{appendix_and_supp}

%% file: appendix_and_supp.tex
\section{Implementation Details}
We provide more details of the implementation of BARON on OV-COCO~\cite{lin2014microsoft} and OV-LVIS~\cite{lvis} benchmarks. 

\myparagraph{Sampling.} For neighborhood sampling strategy, we obtain top $K$ region proposals from the RPN and filter out those with an objectness score lower than 0.85. We also discard regions with an aspect ratio smaller than 0.25 or larger than 4.0. And regions with an area ratio smaller than 0.01 are also discarded. Then we apply NMS on the region proposals with IOU threshold 0.1. The region proposals after NMS are used for neighborhood sampling. We sample $G$ bags of regions for each region proposal with a probability 0.3 to sample each surrounding candidate box. For OV-COCO, we set $K=300$ and $G=3$. For OV-LVIS, we set $K=500$ and $G=4$ due to the denser spatial distribution of object boxes in the LVIS dataset.

\myparagraph{Classification Loss.}We use CE loss as the classification loss $\mathcal{L_\mathrm{cls}}$ on base categories. Given $C$ object categories, we obtain the embedding $f_{i}$ for the name of the $i$-th category by the text encoder ($\mathcal{T}$) of the VLM. We also learn a background embedding for non-object regions. If a region is labeled as the $c$-th category, the classification loss is 

\begin{equation}
    \mathcal{L_\mathrm{cls}} = -\log \frac{\exp(\tau_{\mathrm{cls}} \cdot \langle \mathcal{T}(w), f_c \rangle)}{\sum_{i=0}^{C}\exp(\tau_{\mathrm{cls}} \cdot \langle \mathcal{T}(w), f_i\rangle)},
\end{equation}

\noindent where $\tau_{\mathrm{cls}}$ is the temperature to re-scale the cosine similarity, $f_C$ is the background embedding and $w$ is the embedding (pseudo words) of the region.  On OV-COCO, we set $\tau_{\mathrm{cls}} = 50.0$. And on OV-LVIS, we set $\tau_{\mathrm{cls}} = 100.0$ since there are orders of magnitude more categories defined in the LVIS dataset. 

\myparagraph{Alignment Loss.}
Assuming there are $G$ bags of regions and the image (teacher) and text (student) embeddings for the $k$-th bag of regions are $f_v^{k}$ and $f_t^{k}$, the alignment loss $\mathcal{L_\mathrm{bag}}$ on bag of regions is calculated as 

\begin{equation}
    \mathcal{L}_\mathrm{bag} = -\frac{1}{2} \sum_{k=0}^{G-1}(\log(p_{t, v}^k) + \log(p_{v, t}^k)).
\end{equation}
The $p_{t, v}^k$ and $p_{v, t}^k$ are calculated as \begin{align}
        p_{t, v}^k = \frac{\exp(\tau_{\mathrm{bag}} \cdot \langle f_t^{k}, f_v^{k} \rangle)}{\sum_{l=0}^{G-1}\exp(\tau_{\mathrm{bag}} \cdot \langle f_t^{k}, f_v^{l} \rangle)} \\
         p_{v, t}^k =  \frac{\exp(\tau_{\mathrm{bag}} \cdot \langle f_v^{k}, f_t^{k} \rangle)}{\sum_{l=0}^{G-1}\exp(\tau_{\mathrm{bag}} \cdot \langle f_v^{k}, f_t^{l} \rangle)},
\end{align}
respectively, where $\tau_{\mathrm{bag}}$ is the temperature to re-scale the cosine similarity. Assuming there are totally $N$ regions and the image (teacher) and text (student) embeddings for the $k$-th region are $g_v^{k}$ and $g_t^{k}$, the alignment loss $\mathcal{L_\mathrm{individual}}$ on individual regions is calculated as 

\begin{equation}
    \mathcal{L}_\mathrm{individual} = -\frac{1}{2} \sum_{k=0}^{N-1}(\log(q_{t, v}^k) + \log(q_{v, t}^k)).
\end{equation}
The $q_{t, v}^k$ and $q_{v, t}^k$ are calculated as \begin{align}
        q_{t, v}^k = \frac{\exp(\tau_{\mathrm{individual}} \cdot \langle g_t^{k}, g_v^{k} \rangle)}{\sum_{l=0}^{N-1}\exp(\tau_{\mathrm{individual}} \cdot \langle g_t^{k}, g_v^{l} \rangle)} \\
         q_{v, t}^k =  \frac{\exp(\tau_{\mathrm{individual}} \cdot \langle g_v^{k}, g_t^{k} \rangle)}{\sum_{l=0}^{N-1}\exp(\tau_{\mathrm{individual}} \cdot \langle g_v^{k}, g_t^{l} \rangle)},
\end{align}
respectively, where $\tau_{\mathrm{individual}}$ is the temperature to re-scale the cosine similarity. 

On OV-COCO, we set $\tau_{\mathrm{bag}}=30.0$ and $\tau_{\mathrm{individual}} = 50.0$. Since there are finer-grained definition of categories and denser distribution of object boxes in the LVIS dataset, we set $\tau_{\mathrm{bag}}=20.0$ and $\tau_{\mathrm{individual}} = 30.0$ on OV-LVIS to make the contrastive learning harder.

 \begin{table}[t]
\centering
\caption{
Number of linear layers (\#Layers) mapping region features to pseudo-words
}
 \tablestyle{3pt}{1.1}
\scalebox{1.0}{
\begin{tabular}{c| cc c}
 \midrule
\#Layers & AP$_{50}^{\text{novel}}$ & AP$_{50}^{\text{base}}$ & AP$_{50}$\\
 \midrule
1 & 34.0 &60.4 &53.5\\
2 & 33.9 & 60.5 &53.5 \\
3 & 34.1 & 60.8& 53.8\\
\hline
\end{tabular}
}
\label{tab:layer_num}
\end{table}

\myparagraph{Mapping Region Features to Pseudo-words.} In our implementation, we used a single linear layer to map region features from the detector to pseudo-words. In Table~\ref{tab:layer_num}, we show that adding more linear layers (\#Layers) brings no noticeable improvements. This observation is also in line with Maaz \etal~\cite{maaz2022class} that visual properties can be transferred to language models (LMs) by linearly mapping visual features to the input space of LMs. 

\myparagraph{Random Word Dropout.} As we apply two different supervision to the pseudo words, the training can lead certain words to overfit to certain losses. To alleviate overfitting, we borrow the idea of Dropout~\cite{srivastava2014dropout} in neural networks where neurons are randomly dropped during training to avoid overfitting to specific neurons. We randomly discard pseudo words for each region with a probability $p_{\mathrm{drop}}$. By default, we set $p_{\mathrm{drop}} = 0.5$ for training on both OV-COCO and OV-LVIS.

\myparagraph{Suppression on Novel Categories.}
On OV-COCO, we observe a tendency to overfit on base categories due to the smaller number of categories. And compared with OV-LVIS where the tail categories act as the novel categories, the distribution of novel and base categories on COCO is more balanced. We adopt the following strategies to alleviate suppression on novel categories: (1) detach the objectness prediction branch so that the suppression onto novel categories would not be back-propagated to the backbone; (2) save the sampled region proposals into a cache so that regions covering potential novel categories detected in certain iteration can be preserved throughout the training phase; (3) use the output of the second last layer of the VLM (CLIP) for classification and the final output for aligning bag of regions to reduce the competition between the two types of losses.

\begin{figure*}[ht]
 \centering
    \includegraphics[width=0.99\linewidth]{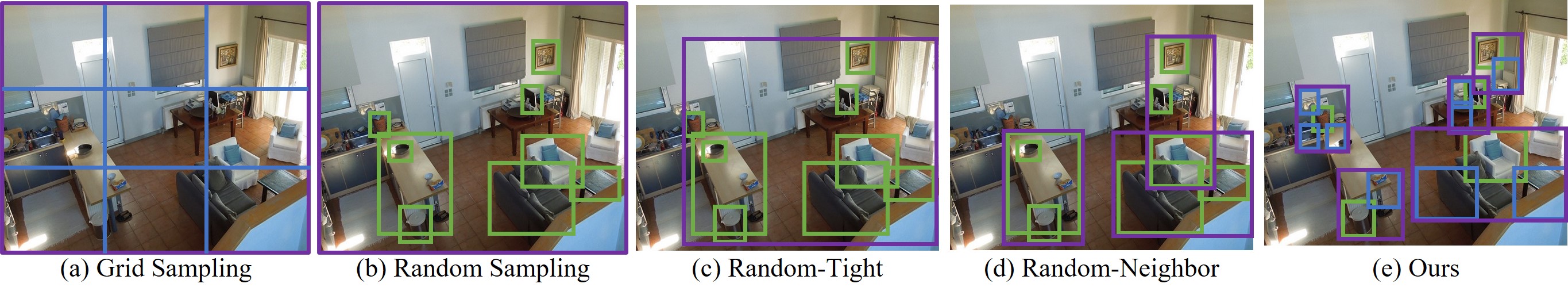}
    \caption{Comparisons of different sampling strategies. Green boxes denote the region proposals. Blue boxes stand for sampled region boxes. The purple box represents the image crop of a bag of regions (a region group).}
    \label{fig:compare_sampling}
\end{figure*}

\begin{table}[t]
\centering
\caption{
Different sampling strategies
}
 \tablestyle{3pt}{1.1}
  \scalebox{1.0}{
  \begin{tabular}{l|c| cc cc}
 \midrule
\#& Strategy & AP$_{50}^{\text{novel}}$ & AP$_{50}^{\text{base}}$ & AP$_{50}$ & \#regions\\
 \midrule
1 & Grid& 25.4 & 58.0 & 49.5&36\\
2 & Random &27.3 & 53.3 &46.5&36\\
3 & Random-Tight & 29.5& 56.9 & 49.7&36\\
4 & Random-Neighbor & 30.7 & 56.9 & 50.0&36\\
5 & Ours (reduced) & \textbf{32.2} & 58.3 & 51.5& 36\\
6 & Ours & \textbf{34.0} & 60.4 & 53.5 & 216 \\
\hline
\end{tabular}
}
\label{tab:supp_sampling_strategies}
\end{table}

\section{Sampling Strategy}

 We have introduced two baseline sampling strategies, \ie. grid sampling and random sampling. The grid sampling strategy is to equally split an image into grids like the pre-training stage in OVR-CNN~\cite{zareian2021open}. And the random sampling strategy is to
randomly sample region proposals to form a bag of regions. These two baseline strategies let the bag of regions represent the whole image. We add two other strategies to shift the focus to neighboring (local) regions.

We start from the random sampling strategy and let the bag of regions represent the image crop that tightly encloses them instead of the whole image (dubbed as Random-Tight). Then we move to the neighborhood centered on region proposals (dubbed as Random-Neighbor). For each center region proposal, we randomly sample 2 nearby region proposals with GIOU larger than 0.5 to make a bag of regions. We randomly take 12 region proposals as centers so that the total number of regions is 36, ensuring a fair comparison with other strategies. Table~\ref{tab:supp_sampling_strategies} shows the performance of these strategies.

In Fig~\ref{fig:compare_sampling}, we show how these sampling strategies differ and how it gradually develops to our final option. In (a), we find the equally split grids may either contain too many objects or only small parts of an object. From (b) to (c), the bag of regions gradually shift to representing neighboring local regions from representing the whole image. However, we observe that there is always box size imbalance such as the left bottom bag of regions in (d). And there are also large area of redundant image contents between the regions in a bag as shown in (c). The box size imbalance and the redundant image contents hinder the image encoder of a VLM to effectively represent a bag of regions. As shown in (e), our sampling strategy obtains a bag of neighboring regions of equal size while capturing potential objects. Although we still observe image contents between sampled regions that do not belong to a bag of regions, they only account for a small portion of the image crop enclosing the bag of regions.

\section{Pseudo Word Encoding}

\begin{figure}[t]
\centering
\includegraphics[width=1.0\linewidth]{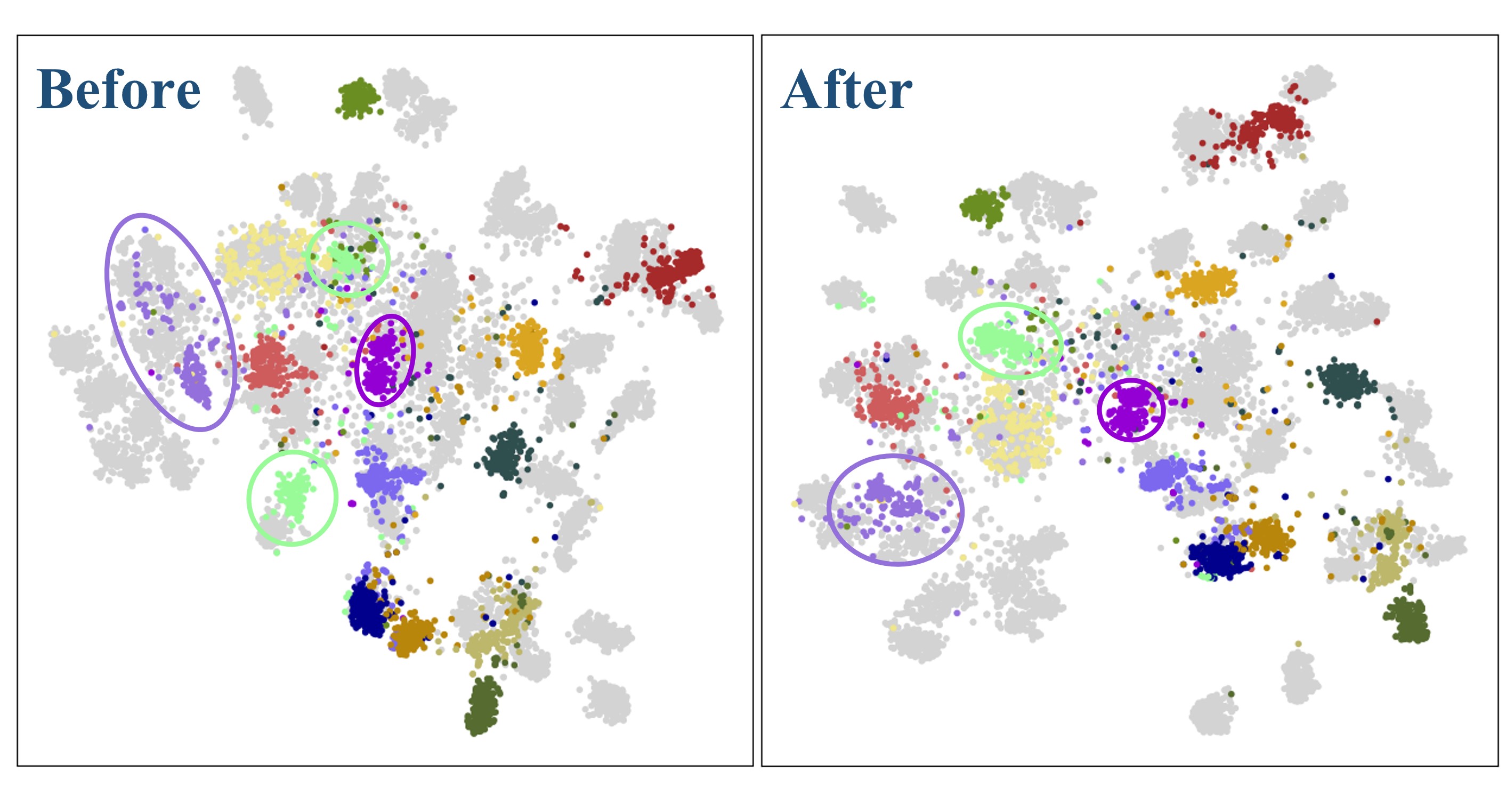}
\caption{tSNE visualization of embeddings on COCO categories. \textbf{Left:} the region features \emph{before} being projected to pseudo words. \textbf{Right:} embeddings \emph{after} sending pseudo words to the text encoder.}
\label{fig:tsne}
\end{figure}

Projecting visual features to word embedding space is common in region-based visual-language representation learning methods~\cite{chen2020uniter,ViLBERT}. In BARON, we project region features into pseudo words to fully exploit the inherent compositional structure of multiple semantic concepts and obtain more distinctive feature embeddings.
In Fig~\ref{fig:tsne}, we show the tSNE visualization of the region features \emph{before} being projected to pseudo words and embeddings \emph{after} sending pseudo words to the text encoder (TE), \ie $\mathcal{T}(w)$. Gray points represent base categories while chromatic points represent novel categories. 
With pseudo words encoded by TE, the categories are split into clusters of a more diverse distribution and distinct boundaries.

\begin{figure*}[ht]
 \centering
    \includegraphics[width=1.0\linewidth]{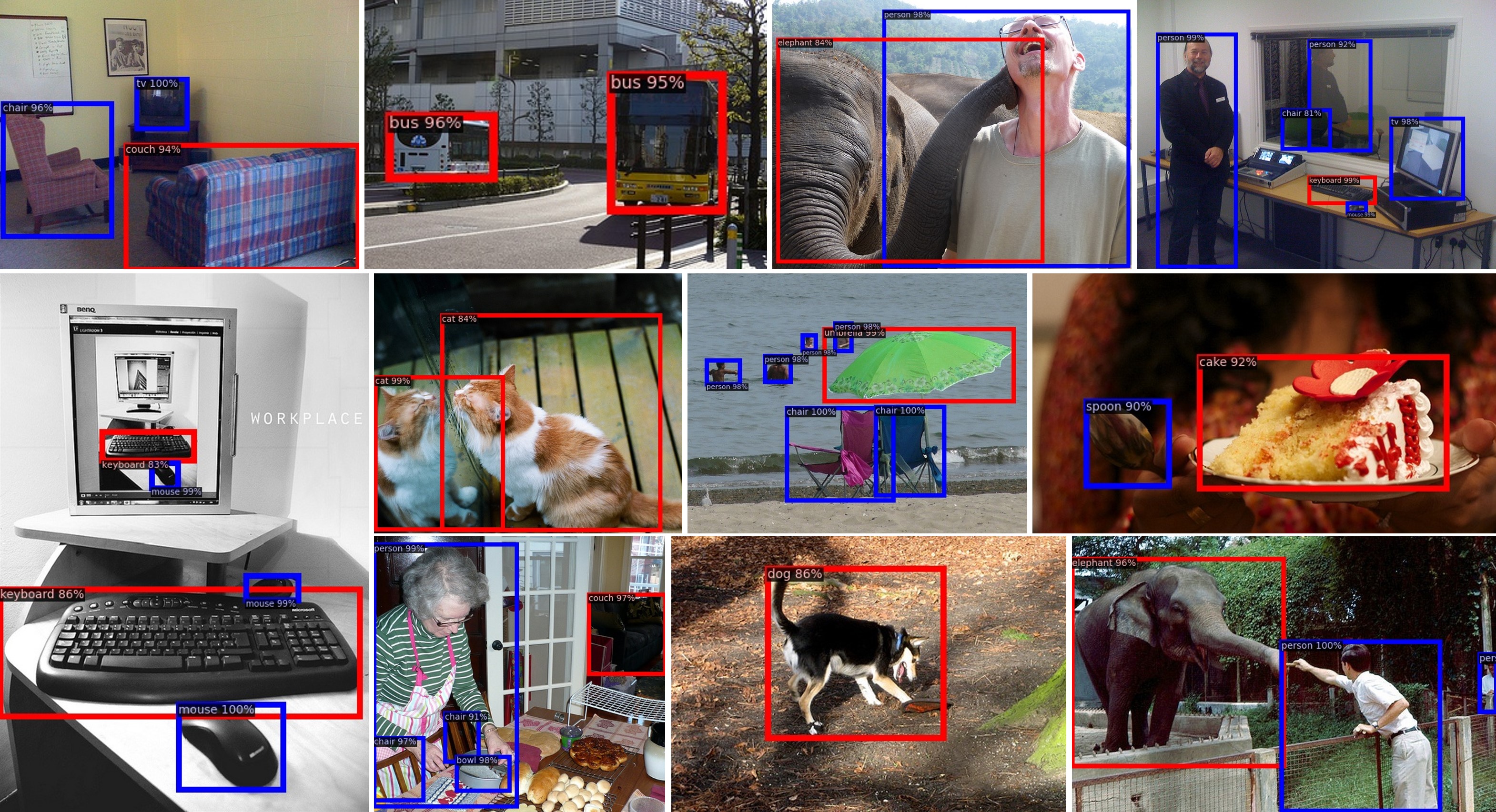}
    \caption{Visualization of detection results on OV-COCO. Red boxes are for novel categories, while blue boxes are for base categories.}
    \label{fig:coco_dets}
\end{figure*}

\section{Image-Guided Inference}
\begin{figure}[t]
\centering
\includegraphics[width=1.0\linewidth]{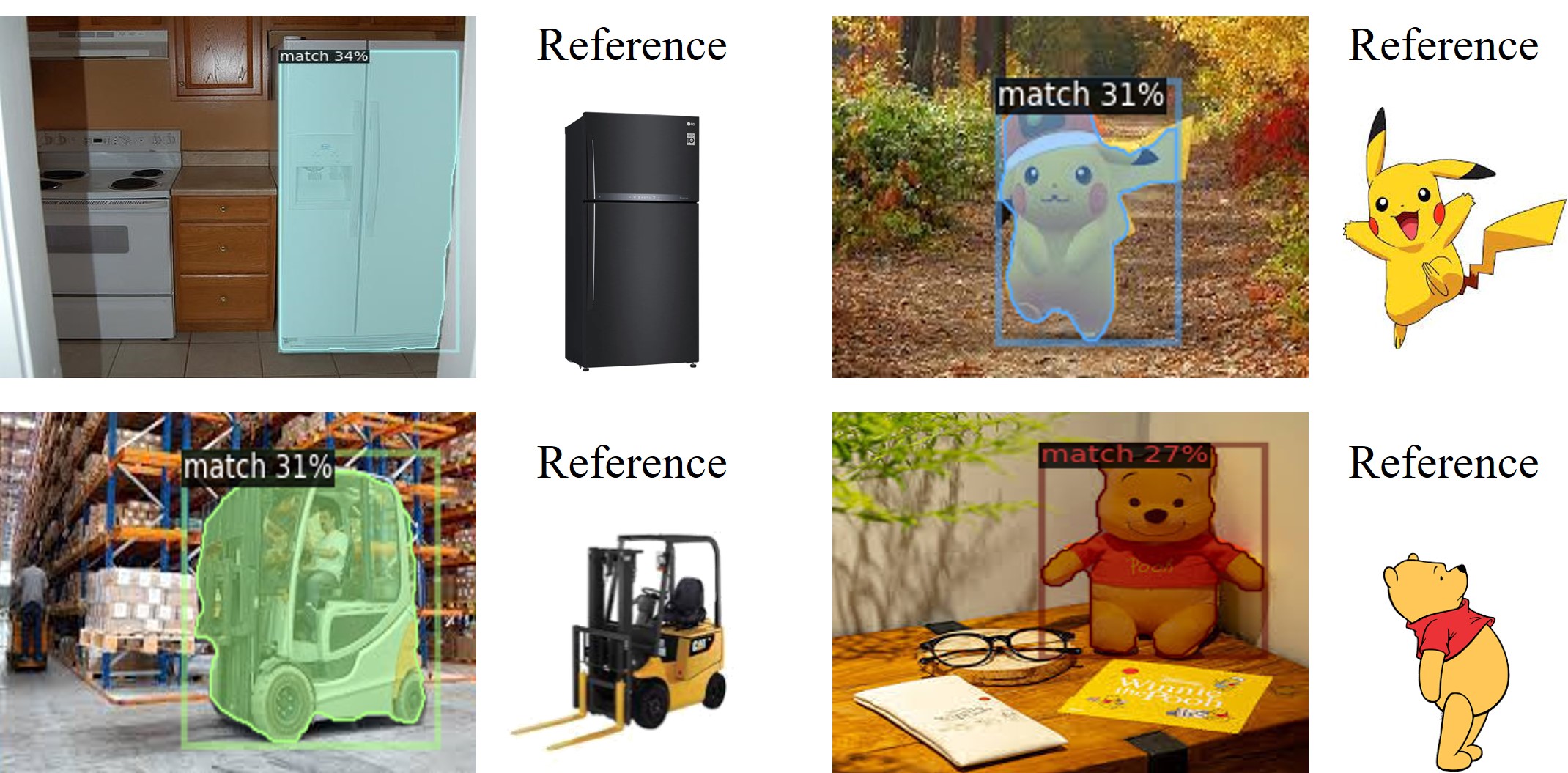}
\caption{Image-guided inference of the detector trained on LVIS dataset. BARON can even recognize the cartoon characters in the reference images (`pikachu' and `winnie pooh').}
\label{fig:ref_images}
\end{figure}

We further examine the generalization ability of our method by using images to guide the inference of the detector. We use the image encoder of CLIP~\cite{radford2021learning} to encode the reference image. And the detector used in this experiment is trained on the LVIS dataset. Given a reference image, our detector is able to detect the object in the reference image as shown in Fig~\ref{fig:ref_images}. Our detector can even recognize the cartoon characters in the reference images (`pikachu' and `winnie pooh').

\section{Detection Results}

\begin{figure*}[ht]
 \centering
    \includegraphics[width=1.0\linewidth]{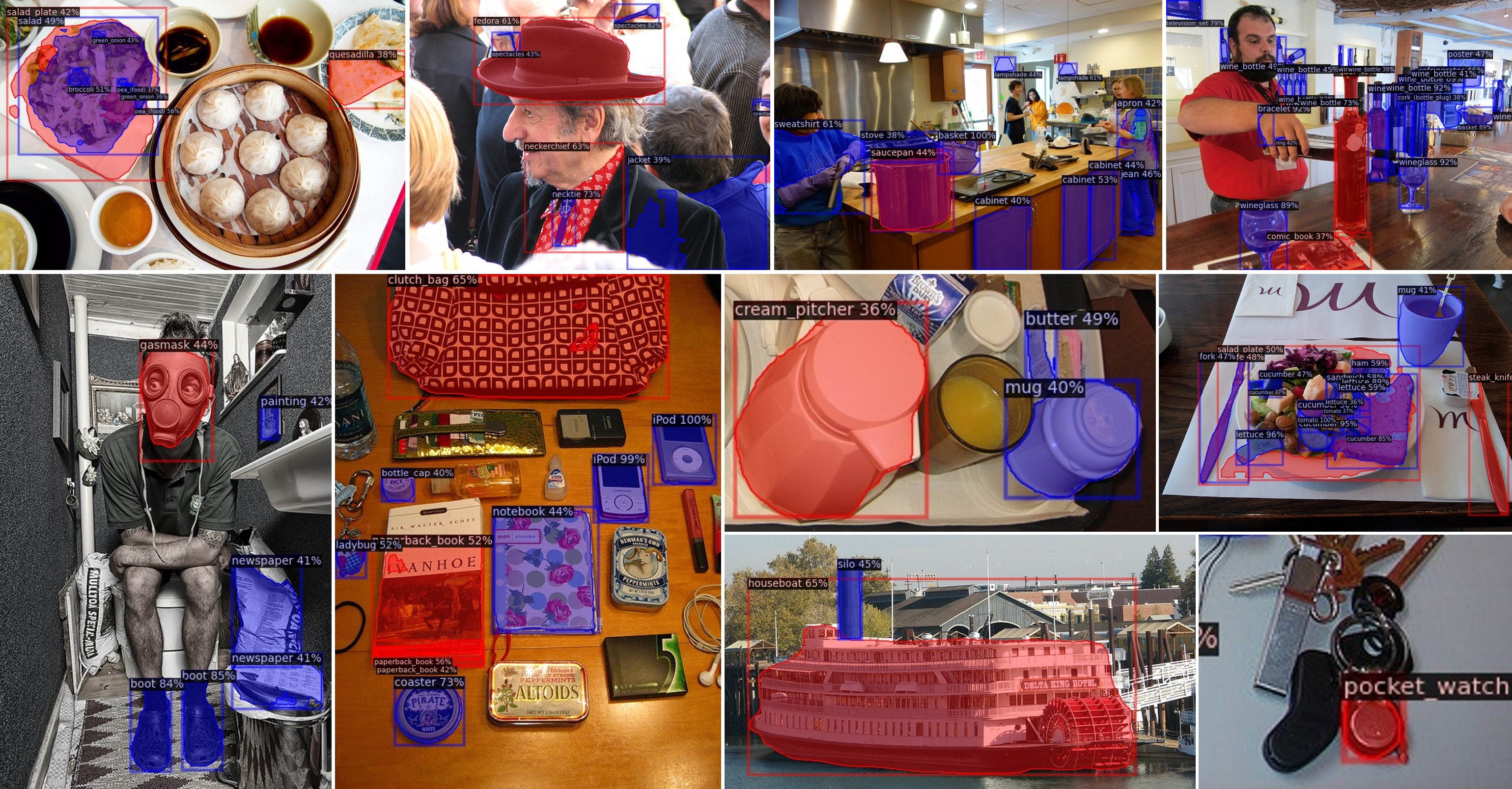}
    \caption{Visualization of detection results on OV-LVIS dataset. Red boxes and masks are for novel (rare) categories, while blue boxes and masks are for base categories.}
    \label{fig:lvis_dets}
\end{figure*}

\begin{figure*}[ht]
\centering
\includegraphics[width=1.0\linewidth]{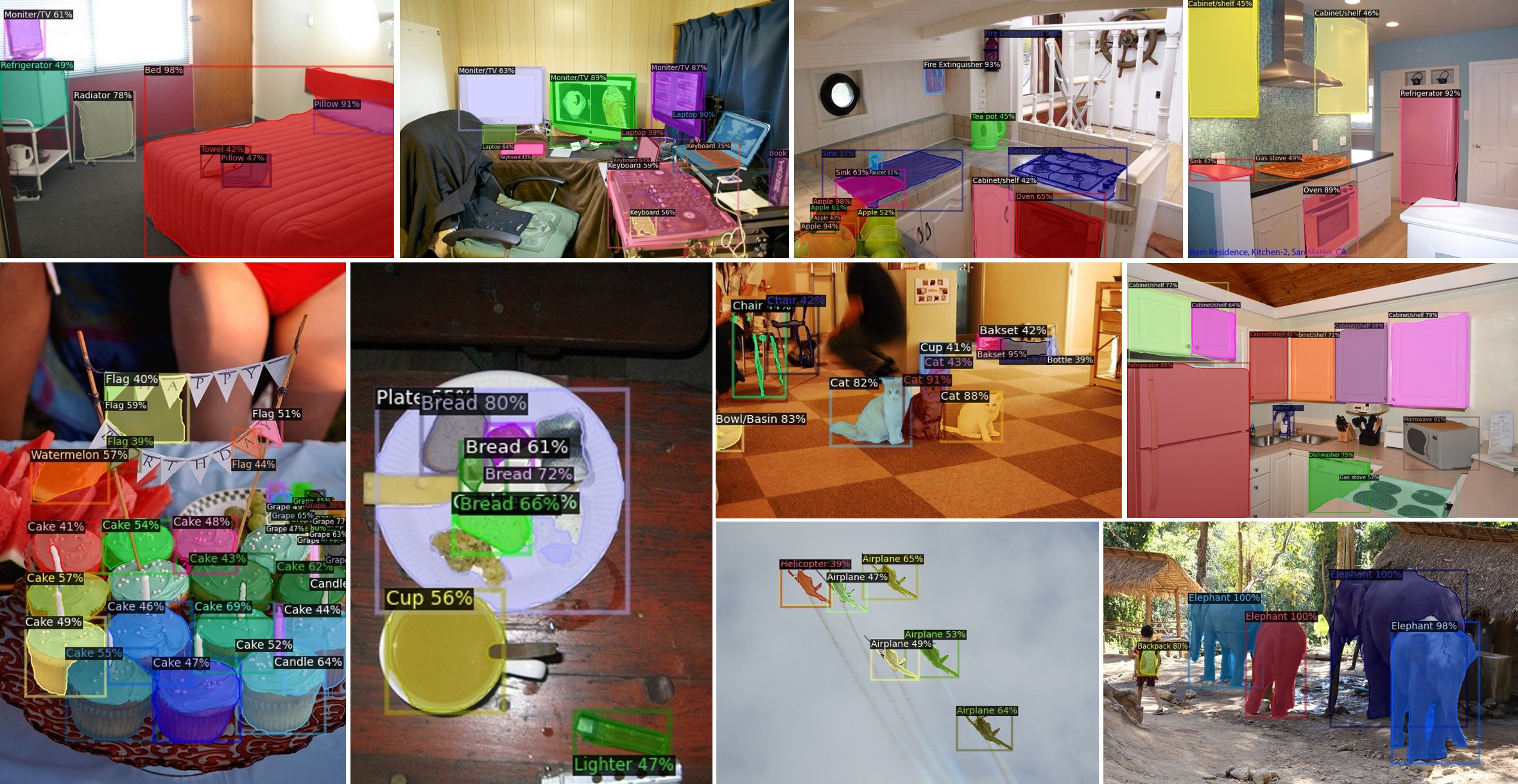}
\caption{Visualization of transfer detection results on Objects365 dataset.}
\label{fig:obj365_dets}
\vspace{30pt}
\end{figure*}

We show more detection results of our method in Fig~\ref{fig:coco_dets} and Fig~\ref{fig:lvis_dets}. On COCO dataset, BARON correctly detects novel categories including bus, keyboard, couch and so on. On LVIS dataset, BARON detects rare categories like salad plate, fedora hat, gas mask and so on. We also visualize the results when transferring the LVIS-trained detector to Objects365~\cite{shao2019objects365} dataset in Fig~\ref{fig:obj365_dets}. We find that the LVIS-trained detector is able to correctly recognize a wide range of object concepts defined in Objects365 dataset, exhibiting impressive generalization ability.

\section{Potential Negative Societal Impacts}
Our models have learned knowledge from vision-language models (VLMs) that are pre-trained on large-scale  web image-text pairs. They potentially inherit and even reinforce harmful biases and stereotypes in the pre-trained VLMs. We suggest scrupulous probing before applying our models for any purpose.

